\documentclass{article}

\usepackage[preprint]{neurips_2024}
\usepackage[table,dvipsnames]{xcolor}
\usepackage{colortbl}
\usepackage{graphicx}
\usepackage{amsmath,amssymb} % define this before the line numbering.
\usepackage{color}

\usepackage{natbib}
\setcitestyle{numbers}
\setcitestyle{square}
\setcitestyle{comma}
\usepackage{hyperref}
\hypersetup{
	colorlinks=true,
	citecolor=teal,
}
\usepackage{url}
\usepackage{booktabs}
\usepackage{microtype}      % microtypography
\usepackage{multicol}
\usepackage{amsmath}
\usepackage{enumitem}
\usepackage{amssymb}
\numberwithin{equation}{section} %for equations in appendix
\usepackage{wrapfig,lipsum,booktabs}
\usepackage{xcolor}
\usepackage{multirow}
\usepackage{float}
\usepackage{graphicx}
\usepackage{subcaption}
\usepackage{algpseudocode}  
\usepackage{algorithmicx,algorithm}
\usepackage{caption}
\usepackage{diagbox} %for supp materials
\usepackage{colortbl}
\usepackage{xcolor}
\definecolor{rowcolor}{rgb}{0.9, 0.9, 0.9}
\usepackage{soul}

\usepackage{algorithmicx,algorithm}
\definecolor{commentcolor}{RGB}{110,154,155}   % define comment color
\definecolor{functioncolor}{RGB}{200,2,127}   % define comment color
\usepackage{array} 
\usepackage{adjustbox}

\newcommand{\ie}[0]{\emph{i.e., }}

\newcommand{\eg}[0]{\emph{e.g., }}

\newcommand{\shortname}{\textit{M$^3$}}
\newcommand{\fullname}{Matryoshka Multimodal Models}
\newcommand{\SpecifiScale}{\textit{SS}}

\title{\fullname{}}

\author{
 Mu Cai$^1$ \hspace{.12in} Jianwei Yang$^2$ \hspace{.12in} Jianfeng Gao$^2$\hspace{.12in} Yong Jae Lee$^1$ \\
   \hspace{0.5cm}\\
  $^1$University of Wisconsin-Madison \hspace{.5in}  $^2$Microsoft Research, Redmond \\
    \hspace{0.5cm}\\
    \url{https://matryoshka-mm.github.io/}
}

\begin{document}

\maketitle

\begin{abstract}
\vspace{-0.1in}
Large Multimodal Models (LMMs) such as LLaVA have shown strong performance in visual-linguistic reasoning.  These models first embed images into a fixed large number of visual tokens and then feed them into a Large Language Model (LLM). However, this design causes an excessive number of tokens for dense visual scenarios such as high-resolution images and videos, leading to great inefficiency. While token pruning and merging methods exist, they produce a single-length output for each image and cannot afford flexibility in trading off information density \textit{v.s.}~efficiency.  Inspired by the concept of Matryoshka Dolls, we propose \textit{\shortname{}: \fullname{}}, which learns to represent visual content as nested sets of visual tokens that capture information across multiple coarse-to-fine granularities. Our approach offers several unique benefits for LMMs: (1) One can explicitly control the visual granularity per test instance during inference, \textit{e.g.}, adjusting the number of tokens used to represent an image based on the anticipated complexity or simplicity of the content; (2) \shortname{} provides a framework for analyzing the granularity needed for existing datasets, where we find that COCO-style benchmarks only need around 9 visual tokens to obtain an accuracy similar to that of using all 576 tokens; (3)
Our approach provides a foundation to explore the best trade-off between performance and visual token length at the sample level, where our investigation reveals that a large gap exists between the oracle upper bound and current fixed-scale representations.

\end{abstract}

\vspace{-10pt}
\section{Introduction}
\vspace{-5pt}

Large Multimodal models~(LMMs)~\cite{GPT4V_System_Card, liu2023llava,zhu2023minigpt, liu2024llavanext, liu2023improvedllava, wang2023cogvlm, Qwen-VL} have shown strong performance in visual-linguistic understanding and reasoning. Models such as LLaVA~\cite{liu2023llava,liu2023improvedllava,liu2024llavanext} 
first embed the input image with a fixed number of visual tokens, and then feed them as prefix tokens to a Large Language Model (LLM)~\cite{Vicuna, llama-3} to reason about the input image. Similar model designs are borrowed in video LMMs~\cite{lin2023video,zhang2023video}, where each frame contributes a fixed number of tokens to form the final video representation.

In reality, the number of visual tokens can be prohibitively large in the case of high-resolution images, and even more so for long videos. Existing works~\cite{lin2023video,liu2024llavanext,zhang2024llavanextvideo,geminiteam2024gemini} mainly tackle this issue by increasing the input context length and consequently, feeding a large number \eg 3-8k of visual tokens into the LLM. This approach has a couple of significant drawbacks: (1) the extremely long context makes both training and inference inefficient; (2) an excessive number of visual tokens can actually \emph{harm} the LMM's performance, distracting it from attending to the relevant information, as we show in Sec.~\ref{sec:exp:video understanding}.  
Several recent works~\cite{bolya2023tome,chen2024image-fastv,shang2024LLaVA-PruMerge} use heuristics to prune and merge visual tokens to reduce the sequence length. However, they produce a single-length output and \emph{do not afford control over the final sequence length}, which could be useful to trade information density versus efficiency while accounting for resource constraints in the deployment phase.

\begin{figure}[t]
\centering
\vspace{-1em}
\includegraphics[width=\linewidth]{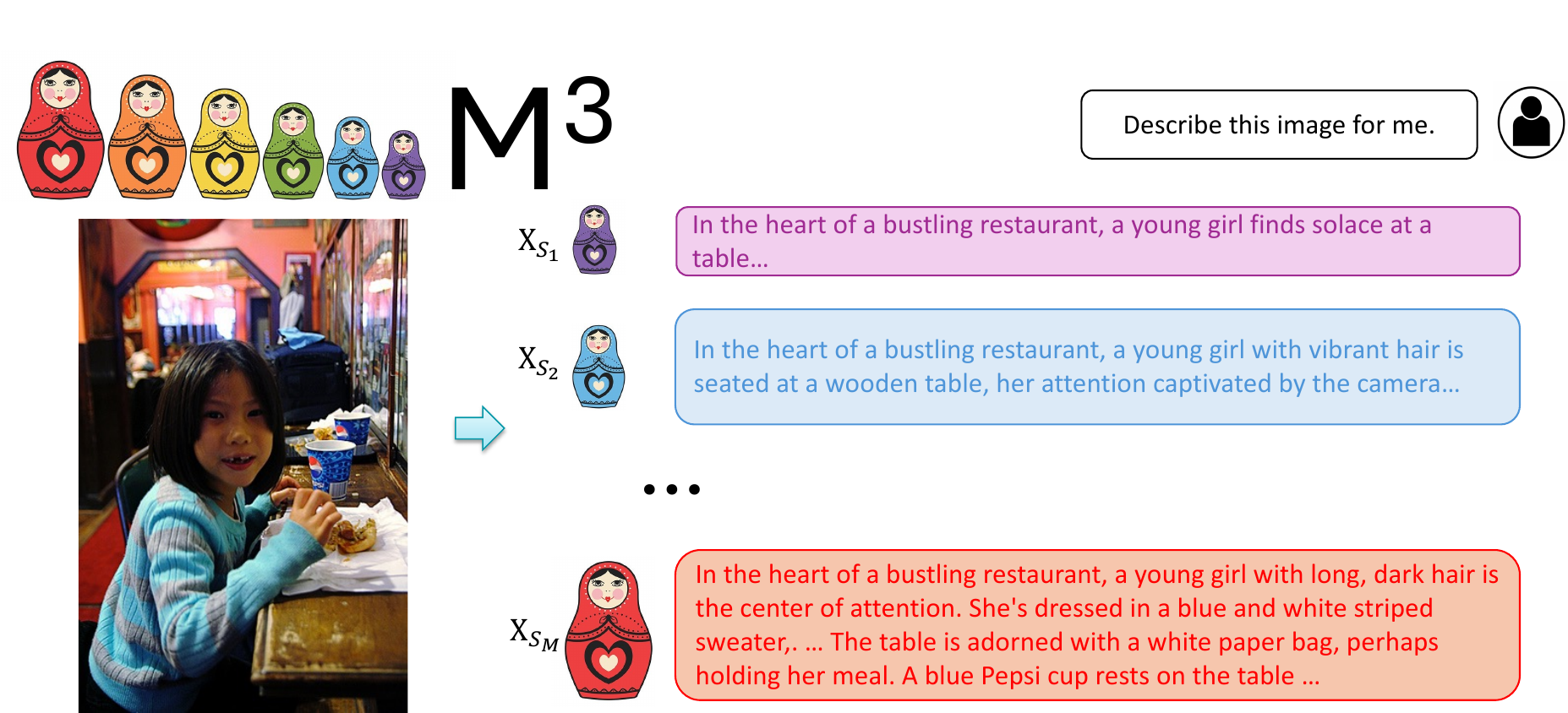}
\caption{
\textbf{\fullname{}.}  We enforce the coarser set of visual tokens $\mathbf{X} _{S_{i-1}}$ to be derived from the finer level of visual tokens $\mathbf{X} _{S_i}$. As a result, the granularity of Matryoshka visual tokens gradually changes in a controllable manner.  The image is from MSCOCO~\cite{lin2014microsoft} validation set.
}

\vspace{-0.1in}
\label{fig:detail-specturm-visualization}
\end{figure}

Images and videos naturally exhibit a hierarchical structure from coarse to fine details, and our human visual system has evolved to recognize visual information in this coarse to fine manner, as shown by biologists and psychologists decades ago~\cite{harris2000coarse,hegde2008time}. Can we create a similar structure for LMMs, where within one suite of model weights, the visual content tokens are organized into different scales of granularities?  Conceptually, our goal is to learn the visual tokens to have a nested structure, similar to the Matryoshka Doll~\cite{kusupati2022matryoshka}. {Matryoshka Representation Learning (MRL)~\cite{kusupati2022matryoshka} builds the Matryoshka mechanism over a neural network's representation vector, where each of the segments with various feature dimensions is capable of handling tasks like classification or retrieval. However, for LMMs, the inefficiency mainly comes from the number of tokens. Thus, inspired by, but different from MRL, our work is motivated to build \fullname{} upon the \emph{token length dimension}, so that we can flexibly adjust it.}

\begin{wrapfigure}{l}{0.5\textwidth} 
  \centering
  \vspace{-10pt}
  \includegraphics[width=0.48\textwidth]{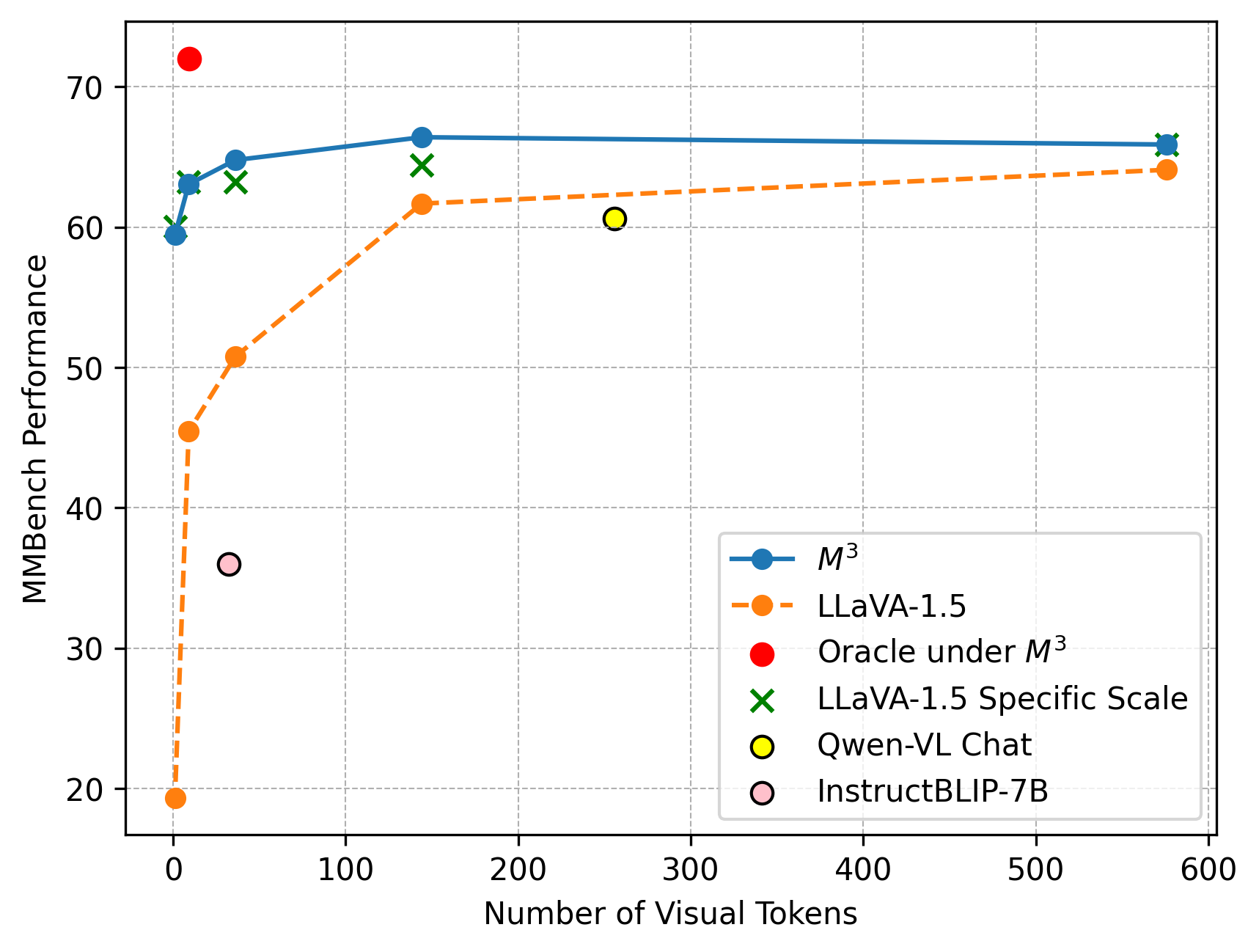}
  \caption{MMBench evaluation results under \shortname{}, oracle under LLaVA-1.5-\shortname{}, LLaVA-1.5 with average pooling at inference time, LLaVA-1.5 separately trained for each specific scale, and other methods. \shortname{} shows as least as good performance as LLaVA trained for each specific scale.  A large gap exists between the oracle upperbound and model's actual performance on a specific scale.}
  \label{fig:oracle curve}
\end{wrapfigure}

Specifically, we propose \textit{\shortname{}: \fullname{}}, which enforces an LMM to learn a hierarchy of visual representation granularities at the token sequence level, instead of the feature dimension level as in MRL~\cite{kusupati2022matryoshka}.  With this representation, at inference time, the visual granularity can be \emph{flexibly controlled} based on specific requirements, e.g., to account for the input image's information density and efficiency constraints.  Our training process is simple and straightforward. During training, we encode the image into $M$ sets of visual tokens from coarse to fine, $\mathbf{X} _{S_i}$, $i = 1, \cdots, M$, where the number of visual tokens gradually increases, \ie $|\mathbf{X}_{S_{i-1}} | < |\mathbf{X}_{S_i}|$.  And importantly, the visual tokens in a coarser level are {derived from} the visual tokens in a finer level, \ie  $\mathbf{X}_{S_{i-1}} \subset \mathbf{X}_{S_i}$, $\forall i$. In this way, the visual information in $ [ {\mathbf{X}} _{S_1}, {\mathbf{X}} _{S_2}, \cdots, {\mathbf{X}} _{S_M}] $ gradually includes more fine-grained details. For example, given a natural image as shown in Figure~\ref{fig:detail-specturm-visualization}, $\mathbf{X} _{S_1}$ includes high-level semantics such as the restaurant and girl, while $\mathbf{X} _{S_M}$  includes more details such as the Pepsi cup and white paper bag.  All other training settings, such as the loss function and model architecture, are kept the same as LLaVA~\cite{liu2023llava,liu2023improvedllava,liu2024llavanext}. 

Our approach, \textit{\shortname{}}, introduces several novel properties and benefits for LMMs.  First, our approach can adaptively and efficiently represent visual content.  Under \textit{one suite of weights}, it generates multiple nested sets of visual tokens with different granualarities in information density.  This enables flexibility in the number of visual tokens used for any image during inference, enabling control over the best tradeoff between cost and performance based on the image or video content. For example, one can use all visual tokens for images with dense details and use just a few tokens for simpler images.  This flexibility can be particularly significant when handling very long visual sequences, such as videos. For instance, given a fixed budget of 2880 visual tokens, a user could represent a video of 2880 frames each with one token or represent the same video by sampling 5 frames each with 576 tokens. 

Second, our approach can be used as a general framework to evaluate the visual complexity of vision-language datasets or benchmarks, \ie which level of granularity is needed in order to perform the given task correctly. {Surprisingly, we find that most benchmarks, especially those mainly crafted from natural scenes (such as COCO)~\cite{goyal2017vqav2,li2023pope,liu2023mmbench}}, can be handled well with only $\sim9$ tokens per image.  In contrast, dense visual perception tasks such as document understanding or OCR~\cite{singh2019textvqa, masry-etal-2022-chartqa} require a greater amount of tokens ($144-576$ tokens) per image to handle the task well. The detailed findings are presented in Sec.~\ref{sec:exp:Image Understanding}.

Finally, our approach provides a foundation to tackle a critical task in LMMs: \textit{How to use the least amount of visual tokens while answering the visual questions correctly?}.  {Based on the model's predictions on the test set, we find that compared to full visual tokens, the oracle can use far fewer tokens while performing much better.}  For example, under six common LMM benchmarks used in LLaVA-NeXT~\cite{liu2024llavanext}, the oracle with the trained~\shortname{} model can use as few as 8.9 visual tokens on average to achieve performance that is 8\% points better than LLaVA-NeXT which uses 576 tokens per image grid. This indicates that there is a large room for improvement compared to the oracle upperbound, as we show in Sec.~\ref{sec:exp:Image Understanding}.

To enable further research on adaptive LMMs that learn diverse information granularities, we publicly release our code and models.

\vspace{-5pt}
\section{Related Work}
\vspace{-5pt}

\textbf{Large Multimodal Models.}
Large Language Models (LLMs) like ChatGPT~\cite{chatgpt}, GPT-4~\cite{gpt4}, and LLaMA~\cite{touvron2023LLaMA} have demonstrated impressive reasoning and generalization capabilities for text. The landscape of LLMs has been significantly transformed by the recent introduction of models that also incorporate visual information, such as GPT-4V(ision)\cite{GPT4V_System_Card}. Building upon open-source LLMs \cite{touvron2023LLaMA, Vicuna}, a plethora of multimodal models have made significant strides, spearheaded by models like LLaVA~\cite{liu2023llava, liu2023improvedllava} and MiniGPT-4~\cite{zhu2023minigpt}, which combine LLaMA's~\cite{touvron2023LLaMA} language capabilities with a CLIP~\cite{radford2021learning} based image encoder. Recently, LMMs on more tasks and modalities have emerged, such as region level LMMs~\cite{cai2024vipllava, zhang2023gpt4roi, chen2023shikra, peng2023kosmos,zhang2023llavagrounding}, 3D LMMs~\cite{3dllm}, and video LMMs~\cite{lin2023video, zhang2023video, zhang2024llavanextvideo}.  However, existing LMMs typically represent the visual content with a large and fixed number of tokens, which makes it challenging to scale to very long visual sequences such as high-resolution images or long-form videos. In this work, we propose to adaptively and efficiently represent the visual content by learning multiple nested sets of visual tokens, providing flexibility in the number of visual tokens used for any image during inference.

\textbf{Matryoshka Representation Learning.} Matryoshka Representation Learning (MRL)~\cite{kusupati2022matryoshka} addresses the need for flexible representations that can adapt to multiple downstream tasks with varying computational resources. This approach, inspired by the nested nature of Matryoshka dolls, encodes information at different granularities within the same high-dimensional feature vector produced by a neural network. The adaptability of MRL extends across different modalities, including vision (ResNet~\cite{he2016deep}, ViT~\cite{dosovitskiy2020vit}), vision + language (ALIGN~\cite{jia2021scaling}), and language (BERT~\cite{devlin2018bert}), demonstrating its versatility and efficiency. Recent work~\cite{li20242d} extends MRL to both the text embedding space and the Transformer layers space. Our approach is inspired by MRL, but instead of learning multiple nested embeddings for a high-dimensional feature vector, we learn \emph{ nested visual tokens  along the token length dimension} for the visual input. 
 We are the first to show that the idea of Matryosha learning can enable explicit control over the visual granularity of the visual content that an LMM processes.

\textbf{Token Reduction.} One of the main causes of inefficiency in recent LMMs is their large number of prefix visual tokens that are fed into the LLM~\cite{liu2023llava,zhu2023minigpt}. The quadratic complexity in Transformers~\cite{vaswani2017attention} is the key issue in scaling the input sequence length for Transformers. Token reduction serves as an effective technique to reduce computational costs in Transformers.  Sparse attention methods such as Linformer~\cite{wang2020linformer} and ReFormer~\cite{kitaev2020reformer} conduct attention operations within local windows rather than the full context, thereby reducing the quadratic complexity of the vanilla attention operation. Another notable method is Token Merging (ToMe)~\cite{bolya2023tome}, which utilizes full attention but gradually reduces the number of tokens in each transformer block by selecting the most representative tokens through bipartite matching {for the Vision Transformer~(ViT). A recent work~\cite{Haurum_2023_ICCVW} further studies different families of token reduction methods for ViT. } However, prior approaches produce a single length output per input image and do not offer multiple granularities over the reduced token sequence. Our \textit{\shortname{}} approach instead learns a multi-granularity, coarse-to-fine token representation within the same model architecture and weights, enabling it to easily be adjusted to various computational or memory constraints.

A concurrent work~\cite{hu2024matryoshka} shares a similar spirit with our approach, representing an image with varying numbers of visual tokens using a single set of model weights. While their method reformats the visual tokens into a sequential list via transformation layers, we use average pooling to preserve the spatial structure of the visual tokens, demonstrating effectiveness in our experiments.
\vspace{-5pt}
\section{\shortname{}: \fullname{}}
\label{sec:approach}
\vspace{-5pt}

\begin{figure}[t]
\centering
\vspace{-1em}
\includegraphics[width=0.99\linewidth]{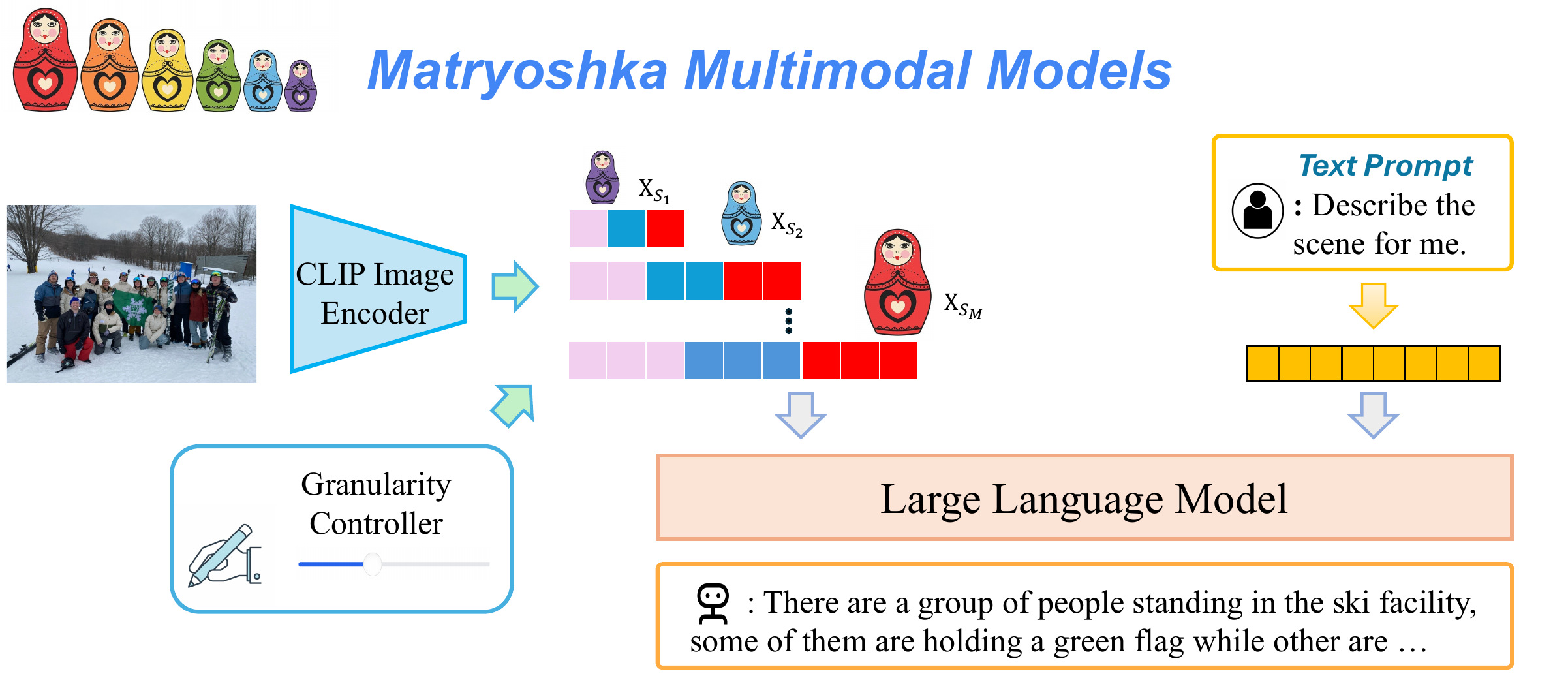}
\caption{
\textbf{Architecture of our proposed \fullname{}.} The visual features from CLIP are represented as several groups of  coarse-to-fine visual tokens. At test time, users can explicitly control the granularity of the visual features. 
}
\vspace{-0.1in}
\label{fig:architecture}
\end{figure}

Our goal is to learn a Large Multimodal Model (LMM) that represents visual content as nested sets of visual tokens capturing information across multiple coarse-to-fine granularities, so that one can explicitly control the visual granularity per test instance during inference. Here we introduce how we learn a Matryoshka doll-like token sequence.

LMMs such as LLaVA~\cite{liu2023llava} typically input a sequence of visual tokens as prefix tokens to the LLM for visual-linguistic reasoning. The visual encoder from pretrained vision-language models, such as CLIP~\cite{radford2021learning} and SigLIP~\cite{zhai2023sigmoid}, is typically utilized to project the images into the set of visual tokens.  
In particular, the CLIP visual encoder represents an input image $\mathbf{I}$ as an $H\times W$ grid of visual tokens $ {\mathbf{X}} _{H\times W} $, where each $\mathbf{X}_i \in \mathbb{R}^{ C}$ is a $C$ dimensional feature vector. Our goal is to learn nested sets of visual tokens $ [ {\mathbf{X}} _{S_1}, {\mathbf{X}} _{S_2}, \cdots, {\mathbf{X}} _{S_M}] $ which encode the visual information in a coarse-to-fine manner. To this end, we enforce ${\mathbf{X}} _{S_i} \subset {\mathbf{X}} _{S_{i+1}}, \forall i$. Importantly, we do not introduce any new learnable parameters to the LMM. We instead optimize the CLIP visual encoder to learn the nested visual representation directly, and train the ensuing LLM to adapt to the learned nested set of tokens. 

For ease of exposition, we consider CLIP-ViT-L-336~\cite{radford2021learning} as the visual encoder, where an image is encoded as $24\times24$ visual tokens (576 total). We create $M$ sets of tokens e.g., $|S_i| \in \{ 1, 9, 36, 144, 576 \}$, in which the visual tokens at the coarser level are derived directly from those at the finer level.  Specifically, given the initial $24\times24$ visual tokens, We sequentially apply $2\times2$ pooling with a stride 2, resulting in $12\times12, 6\times6$, and $3\times3$ visual tokens. Finally, we apply $3\times3$ pooling and get the most condensed single visual token. In this way, the sets of Matryoshka visual tokens can gradually preserve the spatial information in the original tokens while simultaneously forming a coarse-to-fine nested representation.

We train \shortname{} by averaging the autoregressive next token prediction loss for each scale $S_i$ for each image $\mathbf{I}_i$. Specifically, given a Matryoshka visual representation ${\mathbf{X}} _{S_i}$ for scale $S_i$, we maximize the likelihood of the predicted tokens matching the ground-truth answer $\mathbf{X}_{\mathrm{a}}$:
\begin{equation}
P(\mathbf{X}_{\mathrm{a}} \mid {\mathbf{X}}_{S_i}, \mathbf{X}_{\text {q}})=\prod_{j=1}^L P_{\boldsymbol{\theta}}(x_j \mid {\mathbf{X}}_{S_i}, \mathbf{X}_{\text {q}}, \mathbf{X}_{\mathrm{a},<j}),
\end{equation}
where $\boldsymbol{\theta}$ is the trainable parameters of the model, which includes both the CLIP visual encoder and the ensuing LLM. $\mathbf{X}_{\text {q}}$ denotes the question in text format, $L$ denotes the token length of the ground truth answer $\mathbf{X}_{\mathrm{a}}$, and $\mathbf{X}_{\mathrm{a},<j}$ denotes all the ground truth answer tokens before the current prediction token $x_j$, where $j$ denotes the token index during text token generation. We omit system messages for clarity, though they are part of the conditioning. Figure~\ref{fig:architecture} shows our model architecture. 

The final objective averages over all $M$ visual token scales:
\begin{equation}
\min_{\boldsymbol{\theta}} \frac{1}{M} \sum_{i=1}^M -\log P(\mathbf{X}_{\mathrm{a}} \mid {\mathbf{X}}_{S_i}, \mathbf{X}_{\text {q}}).
\end{equation}

With this objective function, \shortname{} learns nested sets of visual tokens that gradually include more details with increasing scale. For example, in Figure~\ref{fig:detail-specturm-visualization}, the smaller set of visual tokens describes the whole scene at a high level while the larger set of visual tokens includes more details such as the Pepsi cup. Our training objective affords our model to conduct visual question answering under any granularity during inference. This can be particularly useful in resource constrained applications; e.g., the visual granularity can be flexibly adjusted based on the anticipated simplicity or complexity of the visual content while taking into account compute and memory constraints.

\vspace{-5pt}
\section{Experiments}
\vspace{-5pt}

In this section, we first detail the experiment settings in Sec~\ref{sec:exp:setting}. Then we show the performance of \shortname{} on both image-level benchmarks~\ref{sec:exp:Image Understanding} and video-level benchmarks~\ref{sec:exp:video understanding}. Finally, we analyze the behavior of \fullname{} and provide ablations in Sec~\ref{sec:exp:analysis} and ~\ref{sec:exp:ablation}.

\subsection{Experiment Settings}
\label{sec:exp:setting}

\vspace{-5pt}\paragraph{Model} We use LLaVA-1.5~\cite{liu2023improvedllava} and LLaVA-NeXT~\cite{liu2024llavanext} as the base LMMs, both with Vicuna 7B as the language model backbone. We finetune the whole model using the exact visual instruction data from LLaVA-1.5 and LLaVA-NeXT, respectively. The learning rate of LLM is $2\times10^{-5}$ and $1\times10^{-5}$, respectively for LLaVA-1.5 and LLaVA-NeXT. The learning rate for the visual encoder is  $2\times10^{-5}$ for both models. We train both models for 1 epoch using 8 NVIDIA H100 GPUs.

Instead of training the language model from scratch, we initialize the language model weights from pre-trained LLaVA-1.5 and LLaVA-NeXT, which we empirically works better. We name our \fullname{} LLaVA-1.5-\shortname{} and LLaVA-NeXT-\shortname{}.

\vspace{-5pt}\paragraph{Visual Token Scales} We design 5 scales for the visual tokens. LLaVA-1.5~\cite{liu2023improvedllava} and LLaVA-NeXT~\cite{liu2024llavanext}  both leverage CLIP-ViT-L-336~\cite{radford2021learning} as the visual encoder, where an image is embedded into $24\times24$ visual tokens. We gradually apply $2\times2$ pooling with stride 2, resulting in $12\times12, 6\times6$, and $3\times3$ visual tokens, where we finally apply a $3\times3$ pooling to get the final single visual token. Therefore, the size of Matryoshka visual token sets are $S \in \{ 1, 9, 36, 144, 576 \}$, following a nested manner. The efficiency anlaysis on the system level is shown in Appendix~\ref{sec: Efficiency Analysis}, where \shortname{} boosts the speed of the LMM prefill process through diminished floating-point operations (FLOPs) and lessens computational memory requirements.

\vspace{-5pt}\paragraph{Evaluations.} For \textbf{image understanding}, we evaluate LLaVA-1.5 and LLaVA-NeXT on (a) diverse multimodal benchmarks: POPE~\cite{li2023pope}, GQA~\cite{hudson2019gqa}, MMBench~\cite{liu2023mmbench}, VizWiz~\cite{gurari2018vizwiz}, SEEDBench~\cite{li2023seed}, ScienceQA~\cite{lu2022learnscienceqa}, MMMU~\cite{yue2023mmmu}, and (b) document understanding/Optical character recognition~(OCR) benchmarks: DocVQA~\cite{mathew2021docvqa}, ChartQA~\cite{masry-etal-2022-chartqa}, AI2D~\cite{ai2d} and TextVQA~\cite{singh2019textvqa}.

For \textbf{video understanding}, we use both (a) open ended video question answering benchmarks evaluated by GPT-3.5: MSVD-QA~\cite{xu2017video}, MSRVTT-QA~\cite{xu2017video} and ActivityNet-QA~\cite{yu2019activitynet}; and (b) multi-choice video question answering benchmarks: NExT-QA~\cite{xiao2021next},  IntentQA~\cite{Li2023IntentQACV}, and EgoSchema~\cite{mangalam2023egoschema}.

\subsection{Image Understanding}
\label{sec:exp:Image Understanding}

\vspace{-5pt}\paragraph{LLaVA-1.5-\shortname{}}
 We evaluate LLaVA-1.5-\shortname{} on the common  multimodal understanding and reasoning benchmarks. Results are shown in Table~\ref{tab:image-level-llava-1.5-ff}. LLaVA-1.5-\shortname{} with full tokens maintains the performance of LLaVA-1.5 across diverse benchmarks. More importantly, our approach shows strong performance even with 1 or 9 tokens.  Specifically, in MMBench, a comprehensive multimodal understanding benchmark, LLaVA-1.5-\shortname{} with 9 tokens surpasses Qwen-VL-Chat with 256 tokens, and achieves similar performance as Qwen-VL-Chat with even 1 token. Compared with InstructBLIP~\cite{instructblip}, LLaVA-1.5\shortname{} with 9 tokens surpasses InstructBLIP-7B and InstructBLIP-13B across all benchmarks. This  demonstrates that our model has both flexibility and strong empirical performance under diverse number of visual tokens.

\begin{table}[t]
    \caption{Comparison between LLaVA-1.5-$M^3$ across various benchmarks under image understanding benchmarks. LLaVA-1.5-\shortname{} maintains  the performance of LLaVA-1.5 while outperforming Qwen-VL and InstructBLIP with fewer tokens. }

    \centering
    \begin{adjustbox}{max width=0.95\textwidth}
    \begin{tabular}{>{\centering\arraybackslash}m{4cm}|c|ccccc}
        \toprule
        {Approach} & {\# Tokens} & {MMBench} & {GQA} & {POPE} & {VizWiz} & {SEEDBench} \\ 
        \midrule
        {Qwen-VL}~\cite{Qwen-VL} & 256 & 38.2 & 59.3 & - & 35.2 & 56.3 \\ 
        {Qwen-VL-Chat}~\cite{Qwen-VL} & 256 & 60.6 & 57.5 & - & 38.9 & 58.2 \\ 
        {InstructBLIP-7B}~\cite{instructblip} &  32 & 36.0 & 49.2 &  - & 34.5 & 53.4 \\
        {InstructBLIP-13B}~\cite{instructblip}  &  32 & - & 49.5 &  78.9 & 33.4 & - \\ 
        \midrule
        {LLaVA-1.5-7B}~\cite{liu2023improvedllava} & 576 & 64.8 & \textbf{62.0} & 85.9 & 54.4 & 60.5 \\ 
        \midrule
        \multirow{5}{*}{LLaVA-1.5-$M^3$} & 576 & 65.9 & 61.9 &\textbf{87.4} & \textbf{54.9} & \textbf{60.6} \\ 
        & 144 & \textbf{66.4} & 61.3 & 87.0 & 53.1 & 59.7 \\
        & 36 & 64.8 & 60.3 & 85.5 & 52.8 & 58.0 \\
        & 9 & 63.1 & 58.0 & 83.4 & 51.9 & 55.4 \\
        & 1 & 59.5 & 52.6 & 78.4 & 49.4 & 50.1 \\
        \bottomrule
    \end{tabular}
    \end{adjustbox}
    % \vspace{1em}
    \vspace{-0.1in}
    \label{tab:image-level-llava-1.5-ff}
\end{table}

\vspace{-5pt}\paragraph{LLaVA-NeXT-\shortname{}}
We use the proposed \fullname{} to finetune LLaVA-NeXT, and compare LLaVA-NeXT-\shortname{} with \SpecifiScale{}, which denotes the setting where the LLaVA-NeXT is trained under a \textbf{S}pecific \textbf{S}cale of visual tokens also for 1 epoch. We also include the oracle upperbound performance. Specifically, `Oracle' denotes the case where the best tradeoff between visual tokens and performance is picked for each test instance. Specifically, for each test instance, we select the the scale with the fewest amount of tokens but can answer the question correctly.  Results are shown in Table~\ref{tab:image-level-LLaVA-NeXT-ff}.  Our approach, \shortname{}, is at least as good as \SpecifiScale{}, while performing better on tasks such as document understanding (TextVQA and ChartQA) and common benchmarks such as MMBench~\cite{liu2023mmbench}.

\begin{table}[t]
    \centering
        \caption{Comparison of approaches with the \SpecifiScale{} baseline and $M^3$ across various benchmarks under LLaVA-NeXT~\cite{liu2024llavanext}. Here {\# Tokens} denotes the number of visual tokens per image grid in LLaVA-NeXT.  \SpecifiScale{} denotes the baseline model trained with a \textbf{S}pecific \textbf{S}cale of visual tokens. \shortname{} is at least as good as \SpecifiScale{}, while performing better on tasks such as TextVQA, ChartQA, and MMBench. \textcolor{blue}{Oracle} denotes the case where the best tradeoff between visual tokens and performance is picked. 
    } 
    \begin{adjustbox}{max width=\textwidth}
    \begin{tabular}{>{\centering\arraybackslash}m{2cm}|c|cccccccc}
        \toprule
     \# Tokens Per Grid & Approach & TextVQA & AI2D & ChartQA & DocVQA & MMBench & POPE & ScienceQA & MMMU \\ 
        \midrule
        \multirow{2}{*}{576} & \SpecifiScale{} & 64.53 & 64.83 & 59.28 & 75.40 & 66.58 & 87.02 & 72.29 & 34.3 \\ 
        & $M^3$ & 63.13 & 66.71 & 58.96 & 72.61 & 67.96 & 87.20 & 72.46 & 34.0 \\ \midrule
        \multirow{2}{*}{144} & \SpecifiScale{} & 62.16 & 65.77 & 55.28 & 67.69 & 67.78 & 87.66 & 72.15 & 36.4 \\
        & $M^3$ & 62.61 & 68.07 & 57.04 & 66.48 & 69.50 & 87.67 & 72.32 & 36.1 \\ \midrule
        \multirow{2}{*}{36} & \SpecifiScale{} & 58.15 & 65.90 & 45.40 & 56.89 & 67.01 & 86.75 & 71.87 & 36.2 \\
        & $M^3$ & 58.71 & 67.36 & 50.24 & 55.94 & 68.56 & 87.29 & 72.11 & 36.8 \\ \midrule
        \multirow{2}{*}{9} & \SpecifiScale{} & 50.95 & 65.06 & 37.76 & 44.21 & 65.29 & 85.62 & 72.37 & 36.8 \\
        & $M^3$ & 51.97 & 66.77 & 42.00 & 43.52 & 67.35 & 86.17 & 71.85 & 35.2 \\ \midrule
        \multirow{2}{*}{1} & \SpecifiScale{} & 38.39 & 63.76 & 28.96 & 33.11 & 61.43 & 82.83 & 72.32 & 35.3 \\
        & $M^3$ & 38.92 & 64.57 & 31.04 & 31.63 & 62.97 & 83.38 & 71.19 & 34.8 \\ \midrule
        \midrule
        \multirow{2}{*}{\textcolor{blue}{Oracle}} & \textcolor{blue}{\# Tokens} & \textcolor{blue}{31.39} & \textcolor{blue}{11.54} & \textcolor{blue}{41.78} & \textcolor{blue}{64.09} & \textcolor{blue}{8.90} & \textcolor{blue}{6.08} & \textcolor{blue}{7.43} & \textcolor{blue}{22.85}  \\
        & \textcolor{blue}{Performance} & \textcolor{blue}{70.51} & \textcolor{blue}{76.36} & \textcolor{blue}{70.76} & \textcolor{blue}{81.73} & \textcolor{blue}{74.35} & \textcolor{blue}{94.29} & \textcolor{blue}{76.07} & \textcolor{blue}{50.44} \\
        \bottomrule
    \end{tabular}
    \end{adjustbox}
    \vspace{-0.1in}
    \label{tab:image-level-LLaVA-NeXT-ff}
\end{table}

Our results also show that dataset level biases towards the visual token scales do exist. For example, ScienceQA maintains consistent performance across all visual token scales. AI2D and MMBench only encounter a small performance drop for even as few as 9 to 1 tokens. On the other hand, dense visual perception tasks such as TextVQA and DocVQA show a significant performance drop with fewer tokens. This analysis shows that \shortname{} could serve as a framework to analyze the granularity that a benchmark needs.

Furthermore, there is a large gap between the model's actual performance under full tokens and the upper-bound oracle. This indicates that using full tokens cannot always result in the optimal performance for all samples; i.e., there is a large room of improvement towards the oracle point.

\subsection{Video Understanding}
\label{sec:exp:video understanding}

Following IG-VLM~\cite{kim2024image}, we directly conduct zero-shot inference on diverse video benchmarks using LLaVA-NeXT-\shortname{}. Specifically, 6 frames are uniformly sampled over the entire video, then arranged as a collage, which is fed into LLaVA-NeXT along with the question to get the response.  Results under  LLaVA-NeXT-\shortname{} and recent video LMMs are show in Table~\ref{tab:LLaVA-NeXT-performance-video}.

LLaVA-NeXT-\shortname{} with full visual tokens again shows comparable performance with LLaVA-NeXT. More interestingly, results indicate that full visual tokens usually \emph{do not lead to the best performance} in video understanding tasks. Specifically, on 4 out of 6 benchmarks, full visual tokens show less desirable performance compared to 720 or 180 visual tokens. We suspect that very long visual context could bring distraction (e.g., too much focus on potentially irrelevant background) to the model's prediction, where a compact representation of the video focusing on the more relevant information may be more advantageous.

Finally, for most video understanding tasks such as ActivityNet, IntentQA and EgoSchema,  with 9 tokens per image grid (45 tokens in total), the accuracy difference compared to full tokens (2880 in total) is less than 1\%. This demonstrates that the video questions in these benchmarks usually require very sparse visual information, as the source of such video understanding benchmarks mostly comes from natural scenes, which matches our observation in image understanding benchmarks.

\begin{table}[t]
    \caption{Overall accuracy of LLaVA-NeXT-\shortname{} and recent video LMMs on various video understanding benchmarks. Here { \# Tokens} denotes the overall number of visual tokens across all frames. }
    \centering
    \begin{adjustbox}{max width=\textwidth}
    \begin{tabular}{>{\centering\arraybackslash}m{4cm}|c|ccc|ccc}
        \toprule
        {Approach} & {\# Tokens} & {MSVD} & {MSRVTT} & {ActivityNet} & {NextQA} & {IntentQA} & {EgoSchema} \\ 
        \midrule

        {Video-LLaMA}~\cite{zhang2023VideoLLAMA} & -  & 51.6 &  29.6  & 12.4  & - & - & -  \\ 
        {LLaMA-Adapter}~\cite{Zhang2023LLaMAAdapterEF} & - & 54.9 & 43.8  & 34.2 & - & - &  -\\ 
        {Video-ChatGPT}~\cite{Maaz2023VideoChatGPTTD}  & -  & 64.9& 49.3  & 35.2 & - & - & - \\ 
        {Video-LLaVA}~\cite{Lin2023VideoLLaVALU}   & 2048& 70.7  & 59.2 & 45.3 & - & - &  -\\ 
        InternVideo~\cite{Wang2022InternVideoGV} &  - & - & - & - & 59.1 & -  & 32.1 \\ 
        
        \midrule
        {LLaVA-NeXT-7B}~\cite{liu2024llavanext} & 2880 & 78.8 & 63.7 & 54.3 & \textbf{63.1} & \textbf{60.3} & 35.8 \\ 
        \midrule
\multirow{5}{*}{LLaVA-NeXT-7B-\shortname{}} & {2880} & 78.2 & \textbf{64.5} & 53.9 & \textbf{63.1 }& 58.8 & 36.8 \\ 
    & {720} & \textbf{79.0} & \textbf{64.5} & \textbf{55.0} & 62.6 & 59.6 & 37.2 \\ 
    & {180} & 77.9 & 63.7 & \textbf{55.0} & 61.4 & 59.3 & 37.6 \\ 
    & {45} & 75.8 & 63.0 & 53.2 & 59.5 & 58.7 & \textbf{38.8} \\ 
    & {5} & 73.5 & 62.7 & 50.8 & 56.5 & 56.7 & 36.2 \\ 
        \bottomrule
    \end{tabular}
    \end{adjustbox}
    \vspace{-0.1in}
    \label{tab:LLaVA-NeXT-performance-video}
\end{table}

\subsection{In-depth Analysis}
\label{sec:exp:analysis}

\vspace{-5pt}\paragraph{\shortname{} shows much stronger performance compared to heuristics based sampling at test time.}

\begin{wrapfigure}{l}{0.5\textwidth} 
\centering
\includegraphics[width=\linewidth]{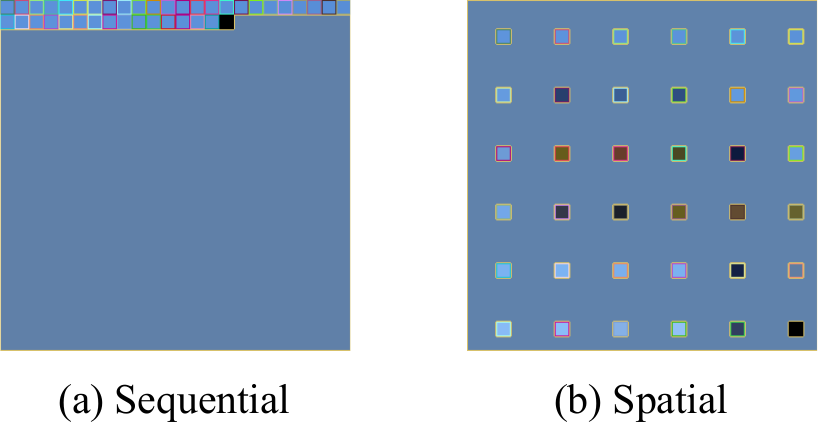}
\vspace{-1em}
\caption{\textbf{Visualization of sequential and spatial sampling.} Given $24\times24$ girds,  the visualized cells denote the sampled tokens.}
\label{fig:vis sampling inference}
\end{wrapfigure}

A simple way to reduce the number of visual tokens via a training-free way is to conduct heuristic token merging or reduction. In Table~\ref{tab:compare inference time sampling}, we compare \shortname{} with three training-free approaches: average pooling, spatial sampling, and sequential sampling. \shortname{} is much more resilient when the number of tokens decreases, while the heuristic based sampling approaches show dramatic performance drop. A visualization of the spatial and sequential sampling is shown in Figure~\ref{fig:vis sampling inference}.  

\begin{table}[t]
    \caption{Comparison between \shortname, and heuristics based sampling baselines---average pooling, spatial sampling, and sequential sampling---at inference time on MMBench with the LLaVA-NeXT architecture.}
    \centering
    \label{tab:compare inference time sampling}
    \begin{adjustbox}{max width=0.9\textwidth}
    \begin{tabular}{>{\centering\arraybackslash}m{2cm}|>{\centering\arraybackslash}m{2cm}|c|c|c}
        \toprule
        \# Tokens & \shortname & Average Pooling & Spatial Sampling & Sequential Sampling \\ 
        \midrule
        576 & 67.96 & 67.18 & 67.18 & 67.18 \\ 
        144 & 69.50 & 61.68 & 65.81 & 60.14 \\ 
        36  & 68.56 & 50.77 & 60.05 & 44.76 \\ 
        9   & 67.35 & 45.45 & 45.45 & 31.96 \\ 
        1   & 62.97 & 19.33 & 26.29 & 22.42 \\ 
        \bottomrule
    \end{tabular}
    \end{adjustbox}
    \vspace{-0.1in}
\end{table}

\begin{figure}
    \centering
\includegraphics[width=0.99\linewidth]{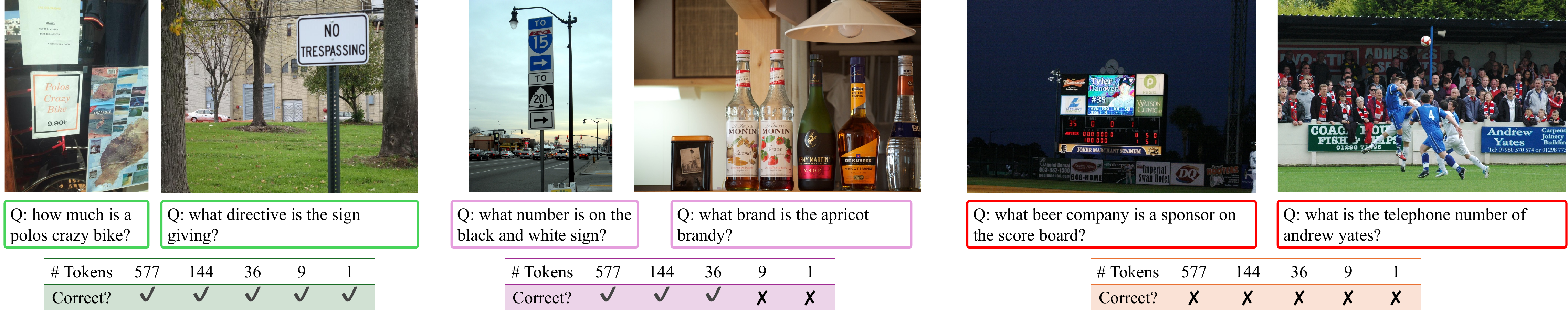}
\caption{
TextVQA test samples with correct and incorrect predictions upon different scales. Answers vary with different number of visual tokens. In addition, \shortname{} can serve as a framework to evaluate the complexity of images.
}
\label{fig:textvqa-visualization}
\end{figure}

\vspace{-5pt}\paragraph{\shortname{} serves as a good metric  for image complexity.} We extract the response from LLaVA-NeXT-\shortname{} in the TextVQA benchmark, and show the samples where using visual tokens across different scales can answer the question correctly and incorrectly.  Shown in Figure~\ref{fig:textvqa-visualization}, the OCR performance aligns with the complexity of the images, which indicates that \shortname{} can be utilized as a metric towards sample level complexity.   

\vspace{-5pt}\paragraph{Large gap between oracle and actual performance.}

As shown in Table~\ref{tab:image-level-LLaVA-NeXT-ff}, the oracle upper-bound can use very few ($6\sim64$) tokens yet achieve at least 10\% better performance compared to full visual tokens. This suggests that a visual token scale predictor, where the model learns to automatically select the best visual token scale given the input images or both input images and questions, has potential to achieve a better tradeoff. This would be interesting future work.

\vspace{-5pt}\paragraph{Zero-shot generalization to longer visual sequences.}

Here we extend the length of the visual tokens at inference time to study the model's zero-shot  generalization behavior. Results under LLaVA-NeXT are shown in Table~\ref{tab:image-grid-performance-generation}. Here LLaVA-NeXT-\shortname{} is trained on  $2\times2$ image grids but evaluated on  $3\times3$ grids. We set the number of visual tokens to be 144 in each image during evaluation. The model obtains a significant improvement in document understanding by 2.12, 1.80, and 4.11 on TextVQA, ChartQA, and DocVQA, respectively, while maintaining the same performance on benchmarks mainly composed of natural scene images. $3\times3$  image grids with 144 tokens per grid own 1440 tokens, yet achieve similar performance with the default LLaVA-NeXT $2\times2$ image grids with 2880 total tokens (576 tokens per grid). This indicates it is promising to feed more subimages while making the number of visual tokens within each subimage much smaller.

\begin{table}[t]
    \centering
    \caption{Performance comparison of different image grid configurations with LLaVA-NeXT-\shortname{}.}
    \begin{adjustbox}{max width=\textwidth}
    \begin{tabular}{c|c|c|ccccccc}
        \toprule
        {\# Grids} & {\# Tokens per grid} & {Overall \# Tokens} & {TextVQA} & {AI2D} & {ChartQA} & {DocVQA} & {MMBench} & {POPE} & {ScienceQA} \\ 
        \midrule
        $2\times2$ & 144 & 720 & 62.61 & 68.07 & 57.04 & 66.48 & 69.50 & 87.67 & 72.32 \\ 
         $3\times3$  & 144 & 1440 & 64.73 & 67.75 & 58.84 & 70.59 & 69.50 & 87.67 & 72.22 \\ 
        \midrule
         $2\times2$   & 576 & 2880 & 63.13 & 66.71 & 58.96 & 72.61 & 67.96 & 87.20 & 72.46 \\ 
        \bottomrule
    \end{tabular}
    \end{adjustbox}
    \vspace{-0.1in}
    \label{tab:image-grid-performance-generation}
\end{table}

\subsection{Ablation Studies}
\label{sec:exp:ablation}

We ablate the key designs in \shortname{}, including the sampling method of Matryoshka visual tokens, and training strategy.

\vspace{-5pt}\paragraph{Matryoshka visual token sampling. } Here we compare three different ways to select the visual tokens for \fullname{}, including average pooling, spatial sampling, and sequential sampling, which is illustrated in Figure~\ref{fig:vis sampling inference}. Shown in Table~\ref{tab:ablation_study_sampling}, averaging pooling shows better performance than the two alternatives across diverse benchmarks. In general, sequential sampling performs the worst. We hypothesize that this is due to the visual tokens having spatial information, while sequential sampling does not naturally align with the spatial distribution of the visual tokens.

\begin{table}[t]
    \centering
    \caption{Ablation on Matryoshka visual token sampling including average pooling, sequential sampling, and spatial sampling.}
    \label{tab:ablation_study_sampling}
    \begin{adjustbox}{max width=\textwidth}
    \begin{tabular}{c|ccc|ccc|ccc}
        \toprule
        & \multicolumn{3}{c|}{TextVQA} & \multicolumn{3}{c|}{MMBench} & \multicolumn{3}{c}{AI2D} \\
        \cmidrule{2-10}
        Num of Vis Tokens & Avg Pooling & Sequential & Spatial & Avg Pooling & Sequential & Spatial & Avg Pooling & Sequential & Spatial \\
        \midrule
        576 & 63.13 & 59.37 & 60.45 & 67.96 & 64.60 & 64.43 & 66.71 & 65.61 & 64.96 \\
        144 & 62.61 & 55.80 & 58.33 & 69.50 & 64.18 & 64.52 & 68.07 & 64.90 & 64.96 \\
        36  & 58.71 & 52.79 & 52.39 & 68.56 & 63.92 & 64.69 & 67.36 & 64.51 & 64.02 \\
        9   & 51.97 & 44.05 & 44.19 & 67.35 & 63.14 & 62.11 & 66.77 & 63.70 & 63.92 \\
        1   & 38.92 & 28.03 & 29.91 & 62.97 & 59.36 & 57.47 & 64.57 & 63.21 & 63.08 \\
        \bottomrule
    \end{tabular}
    \end{adjustbox}
    \vspace{-0.1in}
\end{table}

\begin{table}[t]
    \centering
    \caption{Performance comparison of training LLaVA-NeXT-\shortname{} with and without training the LLM across diverse benchmarks. We see a clear drop when freezing the LLM.}
    \label{tab:ablation train llm}
    \begin{adjustbox}{max width=\textwidth}
    \begin{tabular}{c|cc|cc|cc|cc}
        \toprule
        Num of Vis Tokens & \multicolumn{2}{c|}{TextVQA} & \multicolumn{2}{c|}{MMBench} & \multicolumn{2}{c|}{AI2D} & \multicolumn{2}{c}{DocVQA} \\
        \cmidrule{2-9}
         & w/ LLM & w/o LLM & w/ LLM & w/o LLM & w/ LLM & w/o LLM & w/ LLM & w/o LLM \\
        \midrule
        576 & 63.13 & 61.16 & 67.96 & 63.66 & 66.71 & 63.92 & 72.61 & 69.15 \\
        144 & 62.61 & 57.79 & 69.50 & 65.21 & 68.07 & 63.73 & 66.48 & 59.77 \\
        36  & 58.71 & 49.75 & 68.56 & 63.92 & 67.36 & 62.89 & 55.94 & 44.08 \\
        9   & 51.97 & 36.15 & 67.35 & 61.08 & 66.77 & 62.05 & 43.52 & 28.36 \\
        1   & 38.92 & 19.72 & 62.97 & 51.80 & 64.57 & 60.59 & 31.63 & 17.37 \\
        \bottomrule
    \end{tabular}
    \end{adjustbox}
    \vspace{-0.1in}
\end{table}

\begin{table}[htbp]
    \centering
    \caption{Impact of (a) initializing the LLM weights from LLaVA, and (b) averaging the loss from all scales vs randomly selecting a scale for each sample during training. }
    \begin{adjustbox}{max width=0.95\textwidth}
    \begin{tabular}{c|cccc|cccc}
        \toprule
        {Technique} & \multicolumn{4}{c|}{{TextVQA}} & \multicolumn{4}{c}{{AI2D}} \\
        \midrule
        Init LLM weights from LLaVA  &  & \checkmark &  & \checkmark &  & \checkmark &  & \checkmark \\
        \midrule
        {Average losses over all scales} &  &  & \checkmark & \checkmark &  &  & \checkmark & \checkmark \\
        \midrule
        576 & 60.36 & 62.25 & 61.01 & 63.13 & 62.40 & 65.06 & 65.84 & 66.71 \\
        144 & 59.61 & 61.02 & 59.80 & 62.61 & 63.67 & 65.61 & 65.77 & 68.07 \\
        36 & 54.86 & 55.91 & 55.32 & 58.71 & 63.67 & 65.32 & 66.68 & 67.36 \\
        9 & 46.84 & 47.04 & 48.80 & 51.97 & 63.02 & 64.83 & 65.38 & 66.77 \\
        1 & 33.78 & 33.68 & 36.05 & 38.92 & 61.53 & 63.21 & 63.37 & 64.57 \\
        \bottomrule
    \end{tabular}
    \end{adjustbox}
    \label{tab:ablation smaple or average}
\end{table}

\vspace{-5pt}\paragraph{Training the entire  LMM vs only training CLIP.} Since the nested behavior of Matryoshka visual tokens is learned within the CLIP visual encoder, we next evaluate whether it is necessary to also finetune the LLM. Shown in Table~\ref{tab:ablation train llm}, training the whole LLM achieves better performance. This demonstrates that by also training the LLM, the model can better adapt to the patterns of the visual tokens distributed in the Matryoshka manner.

As explained in Sec.~\ref{sec:approach} and~\ref{sec:exp:setting}, we (a) initialize the LLM weights from LLaVA and (b) minimize the loss averaged upon all visual token scales for each sample during training. An  alternative choice is to randomly sample a visual token scale. Shown in Table~\ref{tab:ablation smaple or average}, initializing the LLM weights from LLaVA and minimizing the losses over all scales shows consistent performance boost compared to using the vanilla text-only pre-trained LLM weights~\cite{Vicuna} and randomly selecting a visual token scale. Initializing the LLM weights from LLaVA makes the training process of \shortname{} more stable.  By learning all scales at once, the model is forced to learn the nested behavior for each sample, which leads to better performance.

\section{Conclusion and Future Work}
\label{sec:conclusion and limitation}

We introduced \shortname{}: \fullname{}, which learns to represent visual content as 
nested sets of visual tokens, capturing information across multiple coarse-to-fine granularities. LMMs equipped with \shortname{} afford explicit control over the visual granularity per test instance during inference. 
We also showed that \shortname{} can serve as an analysis framework to investigate the visual granularity needed for existing datasets, where we discovered that a large number of multimodal benchmarks only need as few as ~9 visual tokens to obtain accuracy similar to that of using all visual tokens, especially for video understanding. Furthermore, we disclosed a large performance-efficiency gap between the oracle upper-bound and the model's performance.

Our work can be naturally extended to other domains. For example, the long context in a text-only LLM or vision tokens in dense vision tasks can also be represented as nested sets of tokens in a Matryoshka manner. One limitation of our current approach is that we are lacking an effective visual token predictor that can bridge the gap between the oracle and LMM's actual performance at a specific scale. We believe this would be an exciting next direction of research in this space.

\section*{Acknowledgement}
This work was supported in part by NSF CAREER IIS2150012, and Institute of Information \& communications Technology Planning \& Evaluation(IITP) grants funded by the Korea government(MSIT) (No. 2022-0-00871, Development of AI Autonomy and Knowledge Enhancement for AI Agent Collaboration) and (No. RS2022-00187238, Development of Large Korean Language Model Technology for Efficient Pre-training), and Microsoft Accelerate Foundation Models Research Program.

\bibliographystyle{unsrt}
\bibliography{main}

\clearpage
\newpage
\appendix

\section{Broader Impact}
\label{sec:boarder_impact}
The broader impact of \shortname{}, a framework with nested visual representations, has potential benefits and risks associated with its deployment and release.  Our model is trained using the exact same architecture and data of LLaVA-1.5~\cite{liu2023improvedllava} and LLaVA-NeXT~\cite{liu2024llavanext}. All the concerns are same as LLaVA. Specifically, as one example, LLaVA conducts instruction tuning using GPT-4 and GPT-4V generated data. The bias from GPT-4 and GPT-4V would still exist in LLaVA.

\section{Efficiency Analysis}
\label{sec: Efficiency Analysis}

To illuminate the computational benefits conferred by \shortname{}, we employ the roofline-based LLM-Viewer analysis as detailed in~\citep{yuan2024llm}. Our analysis is set within a hypothetical context designed to emphasize the effects of \shortname{} on processing efficiency in LMMs. We study the LLaVA-1.5 case where a $336 \times 336$ resolution image is processed using a CLIP-ViT image encoder, resulting in 576 visual tokens. Accompanied by a text prompt with an assumed number of 30 tokens, the nested  visual tokens in \shortname{} substantially lowers the visual token count. The consequences of this reduction are substantial as outlined in Table~\ref{tab:computation Cost}, detailing the computational costs involved in the LMM prefill process. Notably, \shortname{} not only boosts the speed of the LMM prefill process through diminished floating-point operations (FLOPs) but also lessens computational memory requirements.

It is crucial to highlight that the advantages of \shortname{} are not limited to just efficiency improvements. The token reduction approach of \shortname{} can also enhance other LMM acceleration methods, such as quantization and factorization, as referenced in~\citep{yuan2023asvd}. This complementary relationship accentuates the broad potential of \shortname{} to contribute to a wider array of efficiency-boosting strategies.

\begin{table}[h]
\centering
\caption{Computation Cost Analysis. The development device is Tesla V100 GPU, and time estimated by the roofline model represents the theoretical performance that the hardware can achieve.}
\label{tab:computation Cost}
\begin{tabular}{c|c|c|c|c}
\toprule
{\# Tokens} & {FLOPs (TB)} & {Prefill  Time (ms)}  & Total Memory (GB) & Storing Activation (GB) \\ 
\midrule
576                &     8.0                &  58.1     &                21.6     &           3.8         \\ 
144                &   2.2                  &    19.5     &          15.0           &    0.7              \\ 
36                 &    0.9                 &     18.0    &          13.8           &         0.3         \\ 
9                  &       0.5              &     17.7  &           13.6          &        0.2            \\ 
1                  &     0.4                &    17.6   &             13.5        &        0.1            \\ 
\bottomrule
\end{tabular}
\end{table}

\section{More Visualizations on Nested Visual Representation}

\begin{figure}[t]
\centering
\vspace{-1em}
\includegraphics[width=\linewidth]{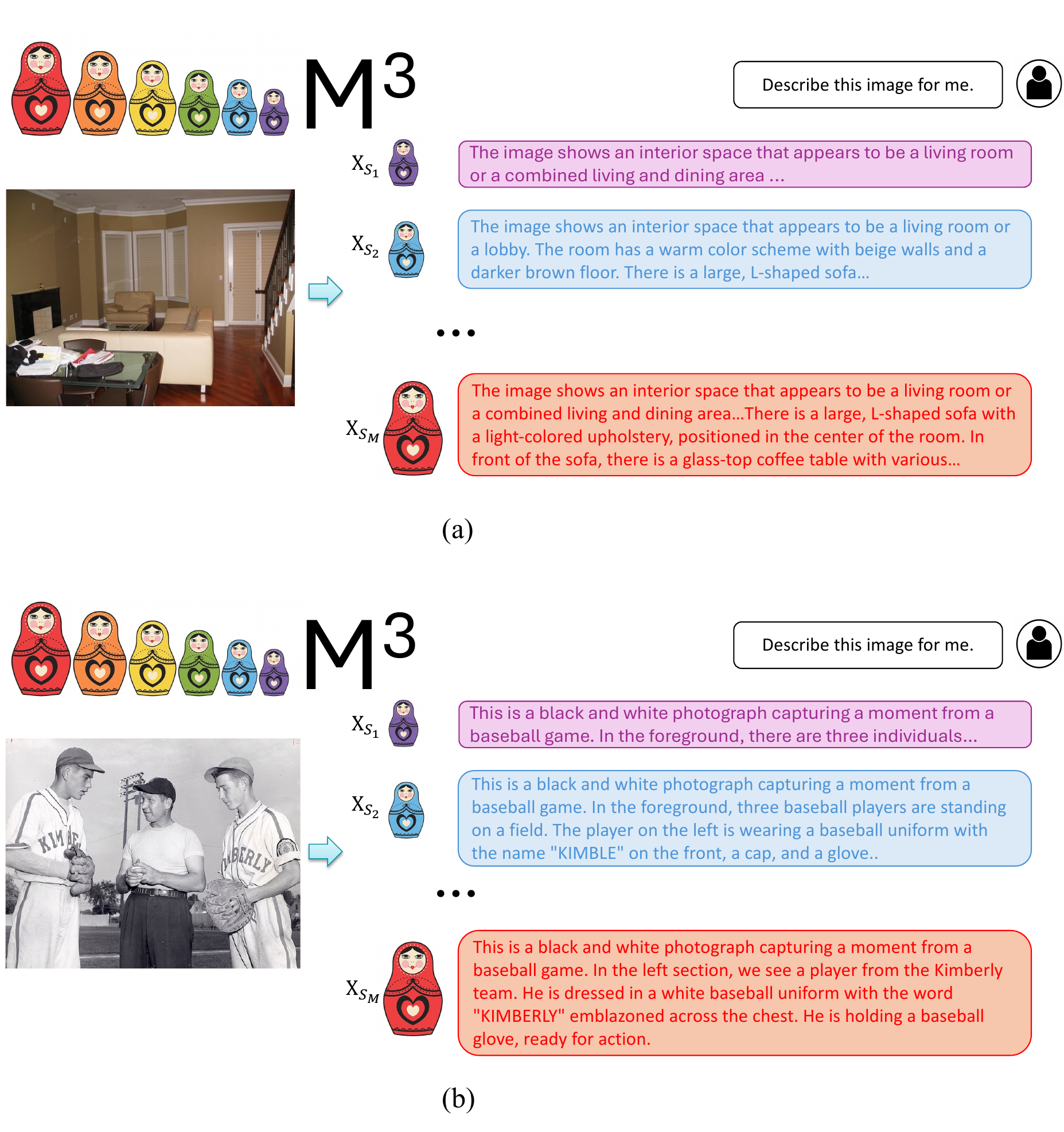}
\caption{
\textbf{More visualization examples.}  With more visual tokens, LMMs can discover more details, and generate higher quality descriptions. The images are from MSCOCO~\cite{lin2014microsoft} validation set.
}
\label{fig:vis-more}
\end{figure}

Shown in Figure~\ref{fig:vis-more}, with more visual tokens, LMMs can discover more details, such as furniture and human attributes.  Besides, LMMs can generate higher quality descriptions with more visual tokens, as demonstrated by the OCR capability in Figure~\ref{fig:vis-more} (b).

\end{document}